\documentclass[10pt,twocolumn,letterpaper]{article}

\usepackage{cvpr}
\usepackage{times}
\usepackage{epsfig}
\usepackage{graphicx}
\usepackage{amsmath}
\usepackage{amssymb}
\usepackage{multirow}
\usepackage{subfigure}

\usepackage[pagebackref=true,breaklinks=true,letterpaper=true,colorlinks,bookmarks=false]{hyperref}

\cvprfinalcopy 

\begin{document}

\title{VSEC-LDA: Boosting Topic Modeling with Embedded Vocabulary Selection}

\author{Yuzhen Ding\\
Arizona State University \\
699 S Mill Ave, Tempe, AZ 85281\\
{\tt\small Yuzhen.Ding@asu.edu}
\and
Baoxin Li\\
Arizona State University\\
699 S Mill Ave, Tempe, AZ 85281\\
{\tt\small Baoxin.Li@asu.edu}
}

\maketitle

\begin{abstract}
Topic modeling has found wide application in many problems where latent structures of the data are crucial for typical inference tasks. When applying a topic model, a relatively standard pre-processing step is to first build a vocabulary of frequent words. Such a general pre-processing step is often independent of the topic modeling stage, and thus there is no guarantee that the pre-generated vocabulary can support the inference of some optimal (or even meaningful) topic models appropriate for a given task, especially for computer vision applications involving ``visual words". In this paper, we propose a new approach to topic modeling, termed Vocabulary-Selection-Embedded Correspondence-LDA (VSEC-LDA), which learns the latent model while simultaneously selecting most relevant words. The selection of words is driven by an entropy-based metric that measures the relative contribution of the words to the underlying model, and is done dynamically while the model is learned. We present three variants of VSEC-LDA and evaluate the proposed approach with experiments on both synthetic and real databases from different applications. The results demonstrate the effectiveness of built-in vocabulary selection and its importance in improving the performance of topic modeling.
\end{abstract}

\section{Introduction}
Topic modeling was originally developed for organizing, interpreting and summarizing large collections of unstructured textual information in document analysis. Typical formulations include Latent Semantic Indexing (LSI) \cite{papadimitriou2000latent}, Probabilistic Latent Semantic Indexing (PLSI) \cite{hofmann1999probabilistic}, Latent Dirichlet Allocation (LDA) \cite{blei2003latent} and some related approaches
\cite{minka2002expectation,griffiths2004finding,buntine2004applying}. Over the years, it has been generalized to various applications with other data types, with different motivations and modified modeling techniques. For example, it can be used for analyzing cross-collection textual data \cite{titov2008modeling,paul2009cross,chen2015differential}, paired text and image data \cite{blei2003modeling,fei2005bayesian,chen2015velda}, and biological or medical data \cite{pritchard2000inference,rubin2016generalized,selvi2019classification}. Nevertheless, the basic idea of topic modeling remains the same, which is to find the hidden structure of the given data by discovering the latent topics. The latent topics are defined as collections/distributions of co-occurring (general-sense) words.

In general, a topic model is a generative hierarchical probabilistic model, and exact inference is intractable for raw data from real applications except for trivial cases. Therefore, in real applications, raw data are typically pre-possessed to create conditioned data with some simplifying properties: real-valued feature vectors may be discretized and represented by indices from a finite set; and a finite vocabulary needs to be determined where each word should have sufficient occurrences in the data to support distribution-based modeling and analysis.

For textual documents, typically a vocabulary is formed by counting the number of occurrences of each word and removing a standard list of stop-words and other words that rarely appear in the corpus \cite{chen2015velda}. For visual documents (images), feature representations such as color histograms, SIFT descriptors \cite{lowe2004distinctive}, etc., are extracted first, then followed by a proper discretization process like $k$-means clustering, to form a raw vocabulary with a reasonable size. The clusters with very few samples may be removed to yield the final vocabulary. As such, the visual words of each image consist of indices corresponding to the final set of $k$-means centers \cite{wang2008spatial,putthividhya2010supervised}. Similar steps are followed for processing biological documents like neuroimaging data, where functional regions are parsed and then represented by a list of terms from which the vocabulary is built \cite{yarkoni2011large}.   

However, such general techniques for pre-processing the raw data are {\sl{}independent} of subsequent topic modeling, which raises the question of whether the discovered topic model is heavily biased by the predefined vocabulary, not reflecting the true latent semantic structures we set out to find. In practice, there are other unanswered questions, like, what is the reasonable size for a vocabulary? what is the optimal set of words to use for a given vocabulary size? etc. Ultimately, it appears that a better solution would be to address the topic modeling and the vocabulary-forming tasks jointly in order to obtain optimal results.

To this end, in this paper we propose a new topic-modeling approach termed Vocabulary-Selection-Embedded  Correspondence-LDA  (VSEC-LDA), which simultaneously addresses the task of latent structure discovery and the task of selecting a suitable vocabulary for supporting model-learning. Both synthetic and real data are used to evaluate the performance of VSEC-LDA. The improved performance of our approach in modeling and processing image-text documents and related tasks not only indicates that a better latent structure can be learned by our model but also shows potential of  our model for diverse applications.

\section{Related Work}
In recent years, a lot of efforts have been made to generalize original topic models to various applications. Many of them focus on modifying the model structure to better fit the specific data  arising from the target applications. Generally, the modification may happen in three different forms: introduction of proper priors or topic-word distributions, utilization of multi-modal topic models, and a combination of the above two.

For the first form, \cite{blei2007correlated} develops the correlated topic model, which replaces the Dirichlet prior with a logistic normal distribution. As an extension, \cite{paisley2012discrete} replaces the Dirichlet prior with a non-parametric setting. Additionally, \cite{blei2010nested} proposes the nested Chinese restaurant process (nCRP) model by introducing an nCRP prior to a tree. Furthermore, \cite{dieng2019dynamic} introduces an additional prior to the topic-word distribution to discover interpretable topics, and \cite{petterson2010word} proposes a words embedding space and uses the cosine similarity between the word and topic vector as the criteria to pick the word which yields better topic cohesion. Some other works \cite{andrzejewski2009incorporating,wang2009decoupling,chen2015differential} introduce constraints to the topic-word distribution so that the model can capture the sharing aspect among the topics and/or the words. The above works have showed their valuable usage in modeling cross-collection textual data. 
For the second form, some attempts have been made for better fitting data from various sources while capturing the correlation among different modalities. Corr-LDA \cite{blei2003modeling} models the conditional correspondence between visual and textual words through discrete indexing that links topics of textual words to visual words. In \cite{putthividhy2010topic}, the model learns a regression from the topics in one modality to those in the other, thus the topic correspondence between different modalities is based on the sets of topics describing each modality rather than one-to-one relationship. The idea has been extended to the scenario where different modalities only have loose correspondence by introducing Markov random field to topic model \cite{jia2011learning}. In a separate line of work, such dual-modal LDA models have been generalized to multi-modal LDA \cite{ahmed2010staying,chen2015velda} to allow learning from multiple views.

While the previous efforts are comprehensive and have provided valuable insights, many approaches also have their limitations. For example, regardless of different model structures, they typically do not consider the construction of vocabulary and modeling at the same time. Instead, for continuous data, they first extract a set of descriptors, then a codebook is created by clustering all the local descriptors using $k$-means, and each descriptor is quantified into a word accordingly. In a separate stage, these discrete words with finite size are fed into the model for learning. Indeed, such a general vocabulary construction pipeline can achieve dimensional reduction, which will alleviate the model workload by quantizing continuous data into discrete format and removing low-frequency descriptors. However, ideally, a model should be able to dynamically select the vocabulary according to its objective. In contrast to previous works, our model incorporates 1) the ability to associate vocabulary construction with model learning, and 2) flexibility in determining the words for each modality. Thus, as we will show, our model not only improves the quality of the estimated topics, but also further reduces computational complexity.

\section{Vocabulary-Selection-Embedded Corr-LDA}
We build our core model VSEC-LDA based on the established Corr-LDA approach \cite{blei2003modeling}, through embedding a dynamic word selection component in latent topic learning. We choose Corr-LDA as the basic building block due to the following two considerations: 1) Corr-LDA has demonstrated performance in many applications with dual modalities and such a dual-modality model can be generalized to multi-modality easily; and 2) Our primary goal is to design an approach embedding vocabulary selection in a typical topic model, not focusing on various topic modeling structures. Due to the page limit, we show the details of two different variants in the supplemental material to further demonstrate the generalization of our method. Our work is based on the assumption that some words in the vocabulary may contribute to the latent model differently than some other words, and possibly some are even detrimental to learning a meaningful topic model, especially if the learning starts from a general-purpose large vocabulary (which is often the case for applications involving visual words built on top of visual features). In this section, we first introduce the architecture of the core VSEC-LDA model, and then describe the word selection and learning process, and at last discuss the tunable threshold in the model. A summary of the notations used throughout the paper is provided in Table \ref{Table1}.
\begin{table}
\centering
\caption{A summary of notations used}
\begin{tabular}{|p{1.6cm}|p{6.4cm}|}
\hline
\multicolumn{2}{|c|}{\textbf{Model specification}}\\
\hline
 symbol & Description \\
 \hline
 \multirow{2}{*}{$w_i$, $v_i$} & The $i$-th text word and visual\\
 &word in the corpus, respectively.\\
 \hline
 $D$ & The number of documents in the corpus.\\
 \hline
 \multirow{2}{*}{$K$ or $E$} & The number of topics in the model.\\
 & $E=0$ in Model I and Model III\\
 \hline
\multirow{3}{*}{$C$ or $S$,$T$ }& The number of unique visual words and\\
& unique textual words,respectively.\\
&$S=0$ in Model I and Model III\\
\hline
\multirow{3}{*}{$R$} & The number of components in each topic\\
& distribution, $R =1$ in Model I and Model II and $R = 2$ in Model III \\
 \hline
 \multirow{2}{*}{$\phi$} &  A $K \times M$ or $E \times M$ matrix indicating\\
 & topic-visual word distribution.\\
 \hline
\multirow{2}{*}{$\theta$} & A $D \times K$ or $D \times E$ matrix indicating\\
& topic multinomial distribution over words.\\
\hline
\multirow{2}{*}{$\psi$} & A $(K+E) \times N$ matrix indicating textual\\
& topic-text word distribution (multinomial).\\
\hline
\multirow{2}{*}{$\pi$}& For Model III. A $R\times(K+E)$ matrix\\
&indicating subregion-topic distribution. \\
\hline
\footnotesize{$\alpha$, $\beta$, $\gamma$, $\eta$, $\delta$} & Model hyper-parameters.\\
\hline
\multicolumn{2}{|c|}{\textbf{Count matrix for model inference}}\\
\hline
$M_{d,c}$, $N_{d,t}$ & \# of $v$ and $w$ in the d-th doc.\\
\hline
$M_{d,k}$ & \# of $v$ in $d^{th}$ doc that are assigned to topic $k$.\\
\hline
$N_{d,k}$ &\# of $w$ in $d^{th}$ doc that are assigned to topic $k$.\\
\hline
$N^{ct}_{k}$, $M^{ct}_{k}$ & the normalized counting matrix used in filtering non-contributing words.\\
\hline
\end{tabular}
\label{Table1}
\end{table}

\subsection{The VSEC-LDA Model}
Figure \ref{fig:model_structure} shows the graphical representation of our proposed VSEC-LDA model (a higher-resolution version is included in supplemental material due to space limitation). In the core model (Model I, shown within the red dash line), each document $d$ is comprised of two modalities: the visual modality consisting of a set of visual words \{\ $v_{1}^{d},v_{2}^{d},...,v_{M_{d}}^{d},v_{f1}^{d},...,v_{IM_{d}}^{d}$ \}\ and the textual modality which consists of a set of textual words \{\ $w_{1}^{d},w_{2}^{d},...,w_{N_{d}}^{d},w_{f1}^{d},...,w_{IN_{d}}^{d}$ \}\  , where $M_{d}$ and $N_{d}$ are the number of relevant visual and textual words in document $d$, respectively; $IM_{d}$ and $IN_{d}$ are the number of irrelevant visual and textual words in document $d$, respectively. Relevance in this context refers to whether the words contribute to the underlying latent topics to be learned. Textual and visual words are correlated in the latent topic space: Each topic of textual words has a corresponding topic of visual words. The generative process of relevant words in our model is similar to that in Corr-LDA, except that the visual words are drawn from a multinomal distribution parametered by $\phi$ instead of a multivariate Gaussian distribution. As for irrelevant words, we assume they are drawn from an uniform distribution (this is for simplicity of discussion; the only real constraint is that they do not contribute to the latent topics in any significant way). We summarize the generative process of each document $d$ in Model I as follows:
\begin{figure}
    \centering
    \includegraphics[width=1\linewidth,height = 6.5cm]{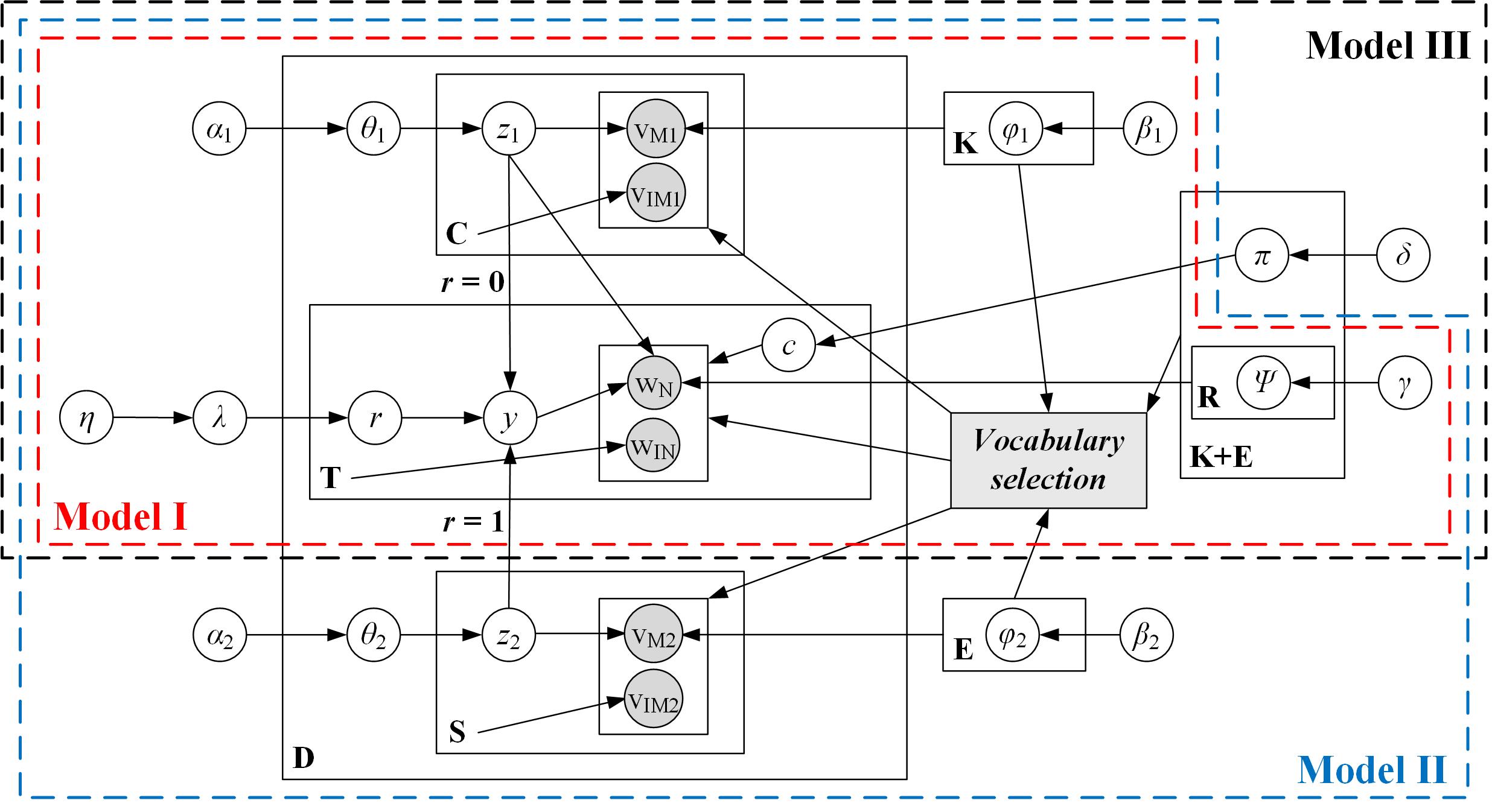}
    \caption{ The generative model in VSEC-LDA. The area within the red dash line is Model I (the core model). The area within the blue dash line represents Model II, which deals with multiple modalities. The area within the black dash line is Model III, targeting neuro-image analysis.}
    \label{fig:model_structure}
\end{figure}
\begin{enumerate}
\item Sample topic proportion $\theta_{d} \sim$ Dirichelet$(\theta_{d}|\alpha)$ for each document $d$.
\item For each relevant visual word $v_m$, $m\in$ \{\ $1,...,M$ \}\
(a) Sample topic indicator variable $z_m$ from Multi( $\theta_{d}$).
(b) Sample the visual word $v_m$ from Multi ($\phi_{z_m}$).

\item For each irrelevant visual word $v_{im}$, ${im}\in$ \{\ $1,...,{IM}_{d}$ \},
sample the visual word $v_{im}$ from Uni( $1,...,{IM}_{d}$).

\item For each relevant textual word $w_n$, $n\in$ \{\ $1,...,N$ \}\

(a) Sample topic indicator variable $y_n$ from Multi ( $\frac{m_{d}^{1}}{m_d},...,\frac{m_{d}^{k}}{m_d},...,\frac{m_{d}^{K}}{m_d}$), where $m_{d}^{k}$ represents the number of times that $k$-th topic appear in current document $d$.
(b) Sample the textual word $w_n$ from Multi ($\psi_{y_n}$).

\item For each irrelevant textual word $w_{in}$, ${in}\in$ \{\ $1,...,{IN}_{d}$\}, sample the textual word $w_{in}$ from Uni( $1,...,IN_{d}$).

\end{enumerate}

Intuitively, from the generative process of this model, it could be learned that for each document $d$, it consists of two different sets of words for both modalities - one set links to some specific topics through topic-word distribution and topic proportion associated with current document $d$, and the other set does not link to any specific topics (i.e., the irrelevant words can be associated with any topics undifferentiated). Therefore, we can update the vocabulary by removing irrelevant words in the learning process based on the distinctive behaviors of the relevant and irrelevant words, and maximize the likelihood of training data regardless of the presentation of the irrelevant words. The joint distribution of the visual / textual words and the latent topics is:
\begin{equation}
\label{eq1}
\begin{split}
p(\bold{v},\bold{w},\bold{z},\theta) & = p(\theta|\alpha)(\prod_{m=1}^{M}p(z_m|\theta)p(v_m|z_m,\phi))) \\
 & (\prod_{n=1}^{N}p(y_n|\bold{v})p(w_n|y_n,\psi)).
\end{split}
\end{equation}
\subsection{Learning the Model}
Since exact inference for the model is intractable, one can rely on one of the usual approximate inference methods, such as variational inference \cite{blei2003latent}, expectation propagation \cite{minka2001expectation}, or Gibbs sampling \cite{griffiths2004finding}. We use Gibbs sampling for its benefit in avoiding local optima while yielding relatively simple algorithms. There are three parameters to be inferred, two topic-word distributions $\phi$ and $\psi$, and one topic proportion for each document $\theta$. These parameters do not need to be directly updated, because they can be integrated out while sampling the topic indicator variables $z_m$ and $y_n$. It is worth emphasizing that the model do not know which words are irrelevant or relevant. Thus, all words are considered in the inference process until after vocabulary selection is conducted.

To initialize the model, first we randomly assign visual topic indicator $z_m$ and textual topic indicator $y_n$ with values drawn from Uniform$(1,...,K)$ for every document $d$. Then, $M_{d,k}$ and $N_{d,k}$ are initialized via counting the number of times that topic $k$ appears in the visual modality and the textual modality, respectively. After that, $M_{d,c}$ and $N_{d,t}$ are filled with the frequency that word $c$ and word $t$ has in current document $d$. We also initialize all the entries in the count matrix used during model inference that have a value zero, by setting them to some small number to avoid numeric stability issues. After initialization, the model iterates through the following steps until convergence:
\begin{enumerate}
\item{Updating hidden topic $\bold{z}$ and $\bold{y}$}

(a) For each visual word $v_m$ in document $d$, update the corresponding visual topic indicator $z_m$ conditioned on 1) the current estimation of topic-visual word distribution $\phi$, 2) the current assignment of topics $\bold{z}_d$, 3) the current estimation of multinomial distribution over topics $\theta_d$, and 4) the current estimation of textual assignment of topics $\bold{y}_d$. 
\begin{equation}
\begin{split}
   &p(z_m = k|\bold{v},\bold{w},\bold{z}_{-m},\bold{y}_d) \\
   & \propto p(z_m = k|\theta_d)p( v_m = c|\phi)p(\bold{y}|z_{-m})\\
   & \propto \frac{M_{d,k,-m}+\alpha_k}{M_d+K\alpha-1}\frac{M_{k,c,-m}+\beta_c}{M_k+C\beta-1}(\frac{M_{d,k}}{M_{d,k,-m}})^{N_{d,k}}.
\end{split}
\end{equation}
(b) For each textual word $w_n$ in document $d$, update the corresponding topic $y_n$, conditioned on 1) the current estimation of topic-textual word distribution $\psi$, 2) the current assignment of visual topics $\bold{z}_d$, and 3) the current assignment of visual topics $\bold{y}_d$.
\begin{equation}
\begin{split}
   &p(y_n = k|\bold{v},\bold{w},\bold{y}_{-n},\bold{z}_d) \\
   &\propto p(y_n = k|\bold{w},\bold{z})p( w_n = t|\psi)\\
   &\propto \frac{M_{d,k}}{M_d}\frac{N_{k,t,-n}+\gamma_t}{N_k+T\gamma-1}.
\end{split}
\end{equation}
(c) Estimate current model parameters by the following equations,
\begin{equation}
\theta_{k,d} = \frac{M_{k,d}+\alpha}{M_d+K\alpha}
\label{Eq:para1}
\end{equation}
\begin{equation}
\phi_{k,c} = \frac{M_{k,c}+\beta}{M_k+C\beta}
\label{Eq:para2}
\end{equation}
\begin{equation}
\psi_{k,t} = \frac{N_{k,t}+\gamma}{N_k+T\gamma}
\label{Eq:para3}
\end{equation}

followed by computing negative log likelihood using Eq. (\ref{Eq:perplexityv}) and Eq. (\ref{Eq:perplexityt}). Note that a lower score indicates better model fitting.
\begin{equation}
    NLL_{c} = exp\{- \frac{\sum_{d = 1}^{D}\sum_{c =1}^{M_{d,c}}\sum_{k=1}^{K}\phi_{k,c} \theta_{k,d}}{\sum_{d=1}^{D}M_{d,c}}\}
    \label{Eq:perplexityv}
\end{equation}
\begin{equation}
    NLL_{t} = exp\{- \frac{\sum_{d = 1}^{D}\sum_{t =1}^{N_{d,t}}\sum_{k=1}^{K}\psi_{k,t} \theta_{k,d}}{\sum_{d=1}^{D}N_{d,t}}\}
    \label{Eq:perplexityt}
\end{equation}

Repeat step (a)-(c) until the negative log-likelihood converges or is lower than the predefined threshold $\zeta$ .

\item{Filtering non-contributing words}

(a) Compute the entropy of each visual word $v_m$ across all the topics. The probability of a certain visual word being picked by a given topic is derived from the current estimation of topic-visual word distribution $\phi$, which can be approximated by the normalized counting matrix $M^{ct}_{k}$ (Eq. (\ref{eq:entropy_v})). Then, words with greater than predefined threshold entropy (the choice of the threshold will be discussed in the next subsection) are removed from the current vocabulary, and $M^{ct}_{k}$, $M_{d,c}$ are updated by deducting the count that is associated with the removed visual words.
\begin{equation}
\label{eq:entropy_v}
\begin{split}
   &Entropy(v_m) \approx -\sum_{k=1}^{K} M^{ct}_{k,m}log(M^{ct}_{k,m})\\  &subject\, to \sum_{k_1}^{K} M^{ct}_{k,m} = 1, \forall \, m \in[1, C]
\end{split}
\end{equation}
(b) Compute the entropy of each textual word $v_m$ across all the topics. The probability of a textual word being picked by a given topic is derived from the current estimation of topic-textual word distribution $\psi$, which can approximate by the normalized counting matrix $N^{ct}_{k}$ (Eq. (\ref{eq:entropy_w})). Then, words with greater than predefined threshold entropy are removed from current vocabulary, and $N^{ct}_{k}$ $N_{d,t}$ are updated by deducting the count that is associated with the removed visual words.
\begin{equation}
\label{eq:entropy_w}
\begin{split}
   &Entropy(w_n) \approx -\sum_{k=1}^{K} N^{ct}_{k,n} log(N^{ct}_{k,n})\\
   &subject\, to \sum_{k_1}^{K} N^{ct}_{k,n} = 1, \forall \, n \in[1, T]
   \end{split}
\end{equation}

\item{Re-estimate hidden topic $\bold{z}$ and $\bold{y}$}

Repeat step 1 and 2 until change of the likelihood is smaller than $\zeta$ and no more word can be filtered. Next, redo step 1 to obtain a better estimation of the hidden topic. This step is not mandatory but was found to be beneficial, since estimating hidden topics and filtering non-contributing words simultaneously might lead to a sub-optimal estimation of the model parameter. Finally, we estimate model parameter using Eq. (\ref{Eq:para1}-\ref{Eq:para3}).
\end{enumerate}

\subsection{Threshold for Vocabulary Selection}
Intuitively, one can rank the entropy of all the words and set the threshold to the point where there is a sharp change (the cut-off point) in the ranked entropy. However, this may have potential issues if the cut-off point happens to be at the tail of the ranked entropy, which indicates the majority of the words will be removed. 
Practically speaking, for any general feature selection and clustering problem, it is reasonable to assume that noise cannot dominate over useful data (or a data-driven learning approach may focus on only the noise). Thus we assume that irrelevant words are the minority and accordingly use this to guide the selection of the cut-off point (e.g., it should not remove more than $30\%$ of the words). Further, in our experiments, it was observed that the cut-off point does not change too much from one iteration to another. Therefore, the threshold is determined from the ranked entropy at the first iteration and it will remain the same throughout the inference process.

Another practical issue is, if we keep removing words according to the threshold, the model may get trapped into the situation where only few words left and thus the partition of words (according to topics) may more likely become disjoint, which is not a desired property for topic models. In our implementation, each time the vocabulary selection is employed, we allow the model to learn without vocabulary selection till some local optimum is reached, and then invoke an iteration of vocabulary selection. Furthermore, in additional to the vocabulary selection threshold, we utilize the threshold $\zeta$ to terminate the training if the model has converged. These steps help to avoid having too few words left. Note that the partition of words primarily depends on the predefined hyper-parameter $\alpha$ of Dirichlet distribution. The parameter $\alpha$ controls the the topic distribution (i.e, how words are related to the topics). The smaller $\alpha$ is, the more disjoint the partition of words may be. Therefore, avoiding using an extremely small $\alpha$ plus the above strategies together will alleviate the concern of having too few words with disjointed topic distributions.

\subsection{Variants of VSEC-LDA}
Model I can be extended to different variants, and we illustrated two here: Model II and Model III, targeting  different applications/purposes. Model II shown in blue dash line in Fig. \ref{fig:model_structure} considers the situation when more than one visual modality is available, and is equivalent to the core model if one of the visual modality is removed. The generative process for each visual modality is the same as that in the core model, and the visual modality is independent of each other. For the textual modality, a relevance variable $r$ is introduced to monitor the relevance between the textual modality and two visual modalities.

Model III (in black dash line in Fig. \ref{fig:model_structure} is for a neuroscience application on analyzing the correspondence between neural activities and brain functions. This model uses GC-LDA as the building block, which employs $R = 2$ components for each topic distribution. This allows the model to learn topics where a single cognitive function is associated with spatially-discontiguous patterns of activation. One can refer to \cite{rubin2016generalized} for more details. With the vocabulary embedded, the text words associated with one topic are more correlated than GC-LDA. 

Due to the page limitation, we will elaborate the two variants in the supplemental material.

\section{Evaluation}

We evaluate our model using both synthetic and real data. Experiments with synthetic data allow us to explicitly check if the model can identify the ideal vocabulary. With real data, we compare Model I to Corr-LDA with different configurations on Corel-5K and SUN \cite{xiao2010sun} databases; Model II is compared with VELDA, a multi-modalities topic model on Wikipedia POTD database; and Model III is compared with GC-LDA on the Neurosynth database. 

\subsection{Experiments with synthetic database}
We use simulations to test if the proposed method can remove non-contributing words in the vocabulary as well as learn the model. The synthetic database used in the simulations is formed by following the generative process of the core model mentioned in the previous section. To be specific, the number of topic $K$ is 20, and the relevant textual and visual words have a size of 80 and 800 respectively. The vocabulary size of irrelevant words in the modalities varies from 20-140 and 200-1400, respectively. We generated 8000 documents for each corpus. For each document $d$,  we first draw topic proportion $\theta_d$ from Dirichlet distribution, then we draw each relevant visual word from multinomial distribution $\phi$. For textual word, first we derive topic proportion given the corresponding visual word, then relevant textual word is drawn from multinomial distribution $\psi$. After relevant part is generated, irrelevant words are drawn from two uniform distributions corresponding to each modality. At last, we randomly select 90\%\ of the corpus as the training set and the remaining 10\%\ as the testing set. Additionally, we fix the model hyper-parameter to be: $\alpha = 0.2$, $\beta = 0.1$ and $\gamma = 0.1$. We set the parameter for Dirichlet distribution to be relatively small, as the experiment with synthetic data only serves as a proof-of-concept. Thus we expect that each document links to some topics, and each topic only picks part of the relevant words with high probability.

We ran the simulation 50 times for each synthetic database. In each experiment, 20 iterations of inference are conducted to yield an initial estimation for all parameters before applying vocabulary selection. Vocabulary selection is performed when the model reaches a local plateau. The experiment automatically terminates when no more words is filtered out and the likelihood stabilizes.

\begin{figure}
  \includegraphics[width=.48\linewidth]{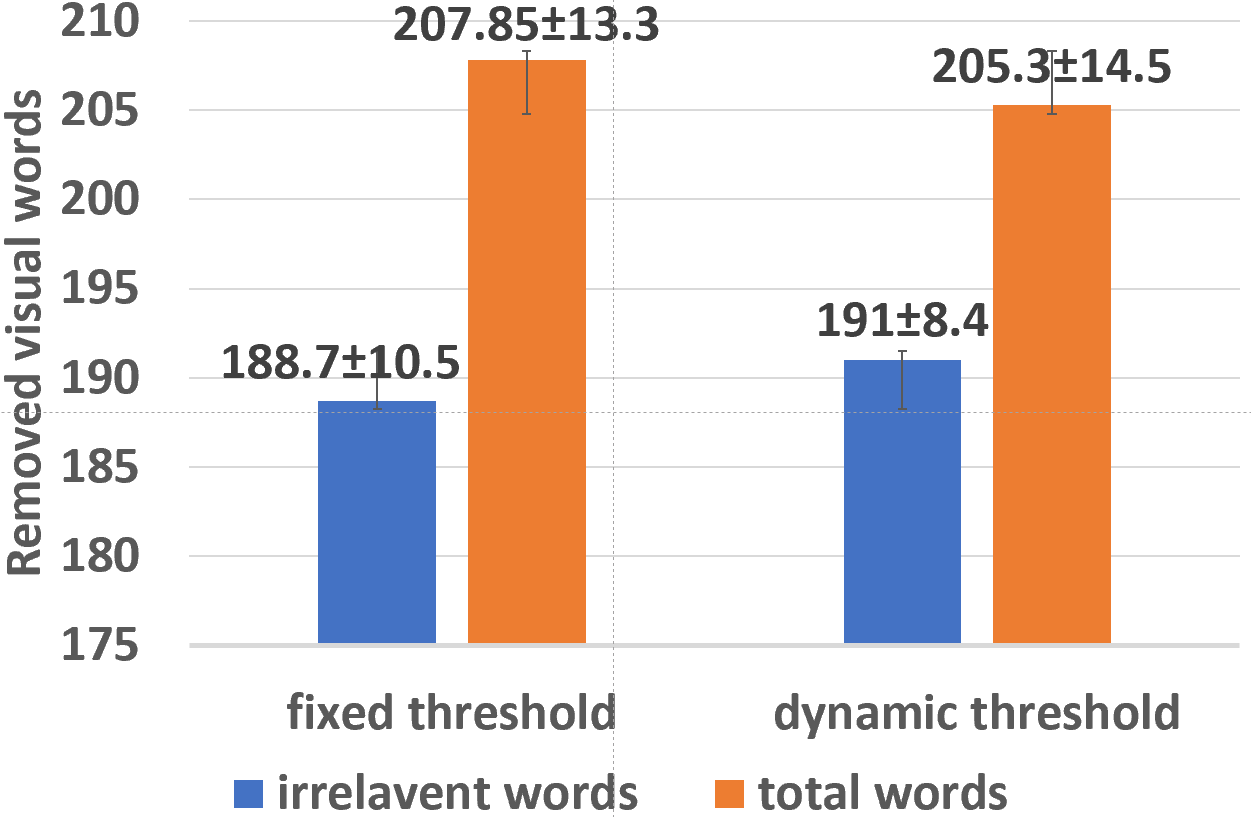}
  \includegraphics[width=.48\linewidth]{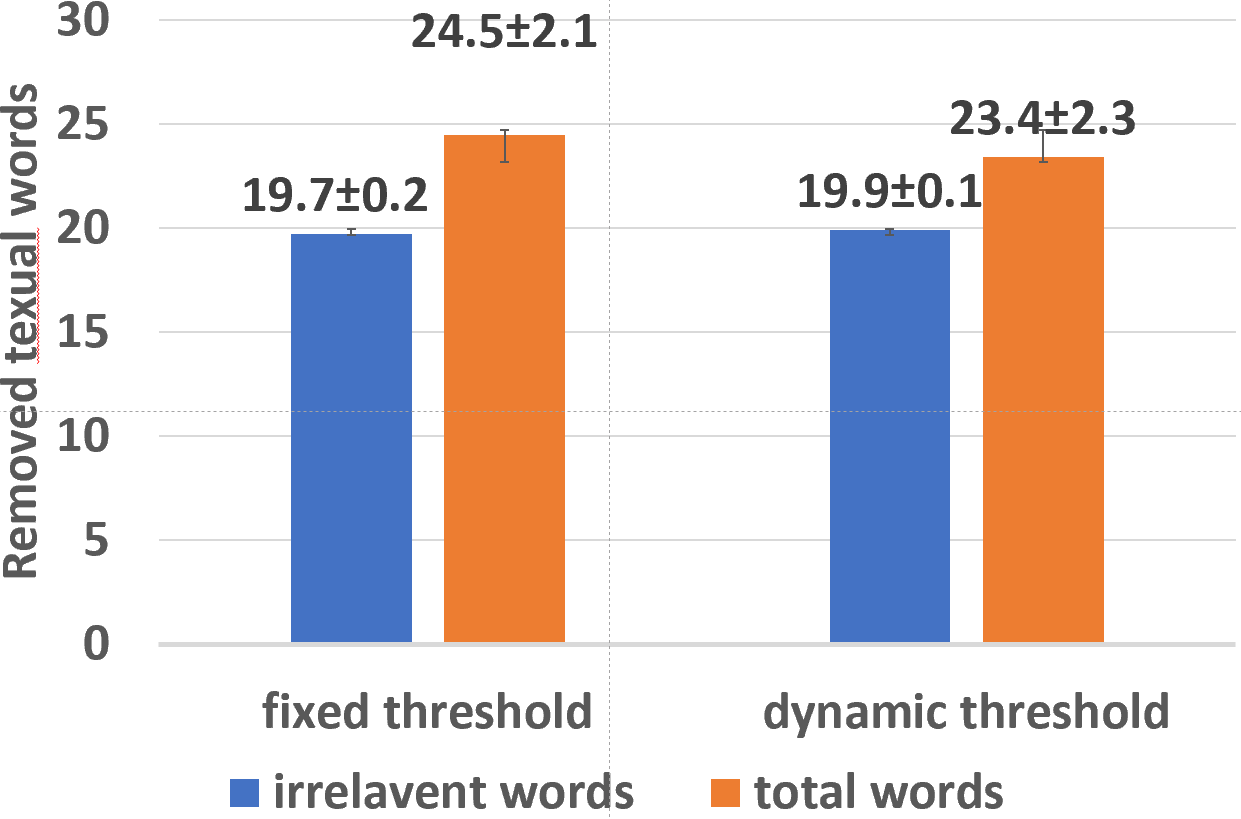}
  \caption{There are 200 irrelevant visual words and 20 irrelevant textual words in the vocabulary of the synthetic database. We show the mean and standard deviation of the number of irrelevant visual / textual words (blue bars) and total visual / textual (orange bars) being removed through 50 rounds of experiments. On average, dynamic threshold yields 3 out of 200 more irrelevant visual words, and 0.2 our of 20 irrelevant textual words being filtered than that of fixed threshold.}
 \label{fig:words_removed}
\end{figure}

Fig. \ref{fig:words_removed} shows the number of filtered irrelevant words with fixed and dynamic filtering threshold for one synthetic database. 
It illustrates that 1) irrelevant words in both visual and textual modalities can be removed by the algorithm and 2) fixed and dynamic thresholds produce similar results. Thus, in the following experiments, we use the fixed threshold to alleviate computational complexity.

\begin{table}
    \centering
    \begin{tabular}{||c||c|c|c||}
    \hline
     irrelevant & 60/600 &100/1000 &140/1400\\\hline
     filtered textual & 59 $\pm$ 2  & 86 $\pm$ 7 &121 $\pm$ 7\\\hline
     filtered visual &587 $\pm$ 11  & 966 $\pm$ 15 & 1121 $\pm$ 15\\ \hline
    \end{tabular}
    \caption{Mean and standard deviation of number of irrelevant words being filtered via 50 rounds experiments. When the number of irrelevant words is greater than 1.5 times of relevant words, it is challenging for the filtering component to do its job within limited iterations.}
    \label{tab:filter_limit}
\end{table}

We also test the limitation of vocabulary selection by experiments with synthetic databases of various number of irrelevant words. We set the maximum training iteration to be 150 and use the fixed filtering threshold for all experiments for fair comparison. Table \ref{tab:filter_limit} shows part of the results given by 50 rounds of experiments for each synthetic database (full results are shown in supplementary material). With the constraint of maximum training iteration, the filtering component starts to see difficulty when irrelevant words become dominant compared to relevant word in the corpus, which is intuitive. Moreover, The trend of average negative log-likelihood, which is used to measure the quality of modeling, also verify the efficiency of our model. The details are presented in the supplementary material.
\subsection{Experiments with Real databases}
Since our model is a multi-modality model, we use Corr-LDA, VELDA, and GC-LDA as baselines, which are all multi-modality models. We first conduct experiments comparing our Model I with Corr-LDA on two real databases with various configurations. Then, we compare the two variants with another two baselines: VELDA and GC-LDA respectively to show the generalization ability of our model.

\subsubsection{Real databases and Experimental Settings}
We use four widely used benchmarks in this paper and the statistic are detailed in the following:

1). \textbf{Corel 5K} contains 4500 and 499 images in the training set and testing set,respectively. Each image is manually annotated with 1 to 5 tags. The word vocabulary contains 260 words. 2). \textbf{SUN} is a large scene understanding database. We build a balanced subset by selecting images from four categories with 4,749 images in total, and around 1,000 images in each category. We have 647 words in the textual vocabulary. 3). \textbf{Wikipedia POTD} is a set of pictures with descriptions from Wikipedia. We use the database provided by \cite{chen2015velda}, which has 2,274 documents for training and 250 for testing. There are 1,000 and 997 words for two visual modalities respectively. The textual modality has 3224 words. 4). \textbf{Neurosynth} is a publicly available database consisting of data automatically extracted from a large collection of functional magnetic resonance imaging (fMRI) publications. The database employed here has 11,362 total publications, which has on average 35 peak activation tokens (viewed as a visual modality), and 46 word tokens (viewed as a textual modality) after preprocessing.


\begin{figure*}
    \centering
    \includegraphics[width=0.32\linewidth,height = 2.5cm]{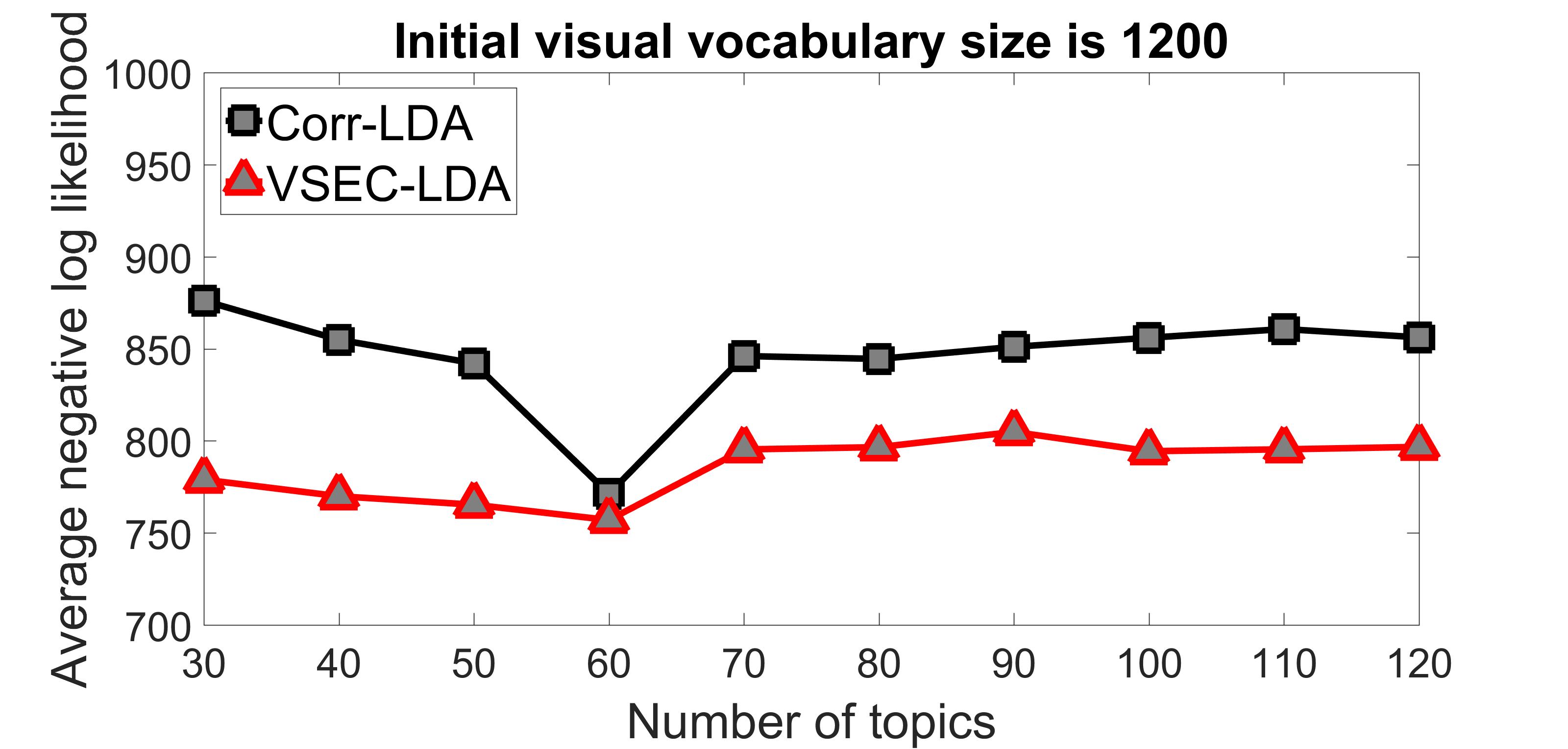}
    \includegraphics[width=0.32\linewidth,height = 2.5cm]{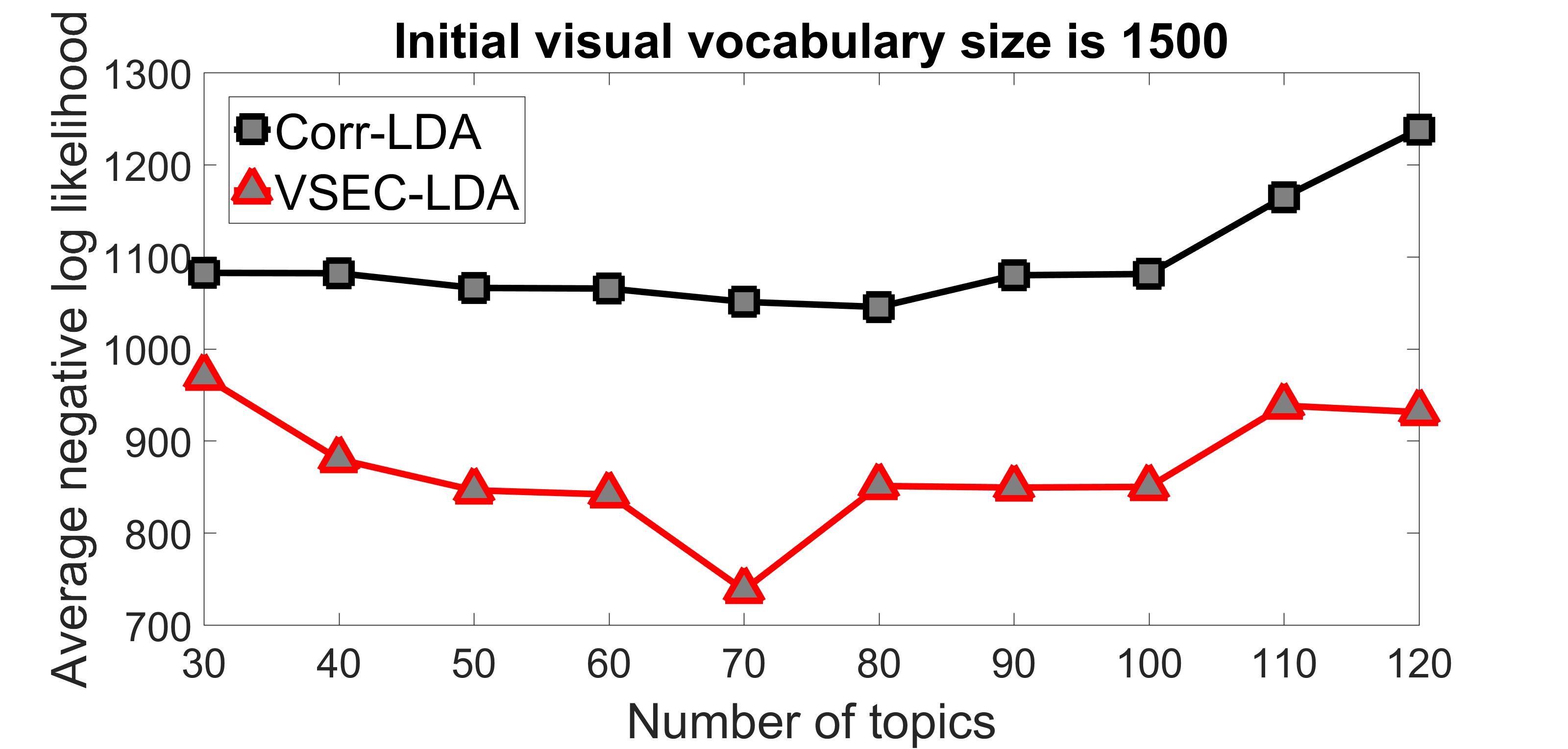}
    \includegraphics[width=0.32\linewidth,height = 2.5cm]{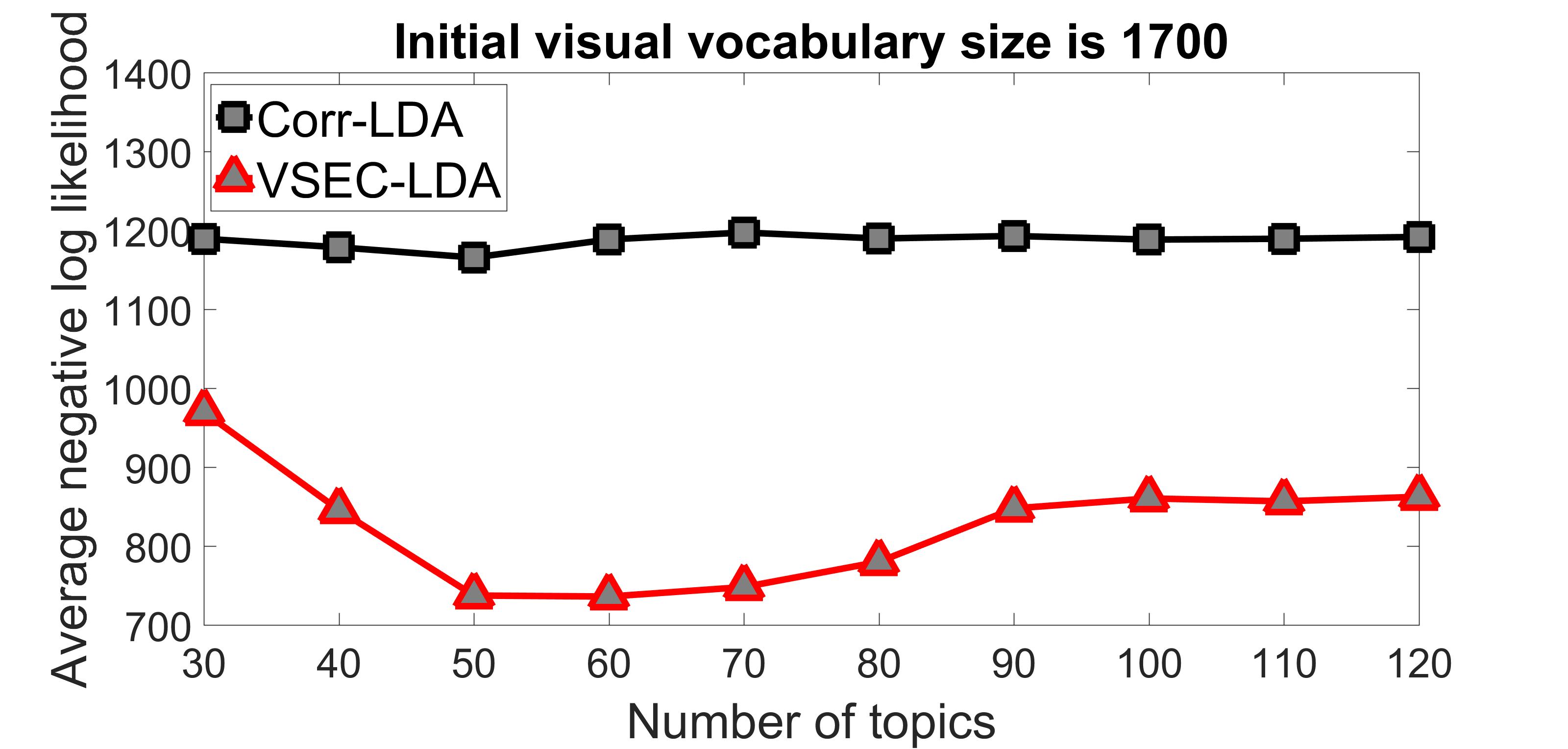}
    \caption{The average held-out likelihood on Corel-5K along with different number of topics $k$ of Corr-LDA and VSEC-LDA. From left to right, the visual vocabulary size range from 1200 to 1700, while the textual vocabulary size is fixed at 260. The average held-out likelihood of both methods increases along with the growing vocabulary. For our method, we achieve similar optimal likelihood values with visual vocabulary of different initial size.}
    \label{fig:testsetlikelihood}
\end{figure*}

We follow \cite{fei2005bayesian,wang2008spatial} to restrict visual vocabulary to SIFT descriptors. We build three visual vocabularies of size 1,200, 1,500 and 1,700 for Corel-5K and 1,000, 1,500 and 2,000 for SUN, respectively. We use the Wikipedia POTD and Neurosynth database provided by the author without further preprocessing. For fair comparison, we fix the hyper-parameters to typical settings: $\alpha = 1$, $\beta = 0.1$ and $\gamma = 0.1$ and tune the number of topics $K$ for Model I and Corr-LDA. We test both methods with different settings: 1)filtering both modalities/ one modality, 2) doing/not doing re-estimation of hyper-parameters. In the comparison experiments with VELDA and GC-LDA, we simply following their settings for fairness.

\subsubsection{Results and Analysis}
The textual vocabulary is built with tags/objects, which are typically from a pre-processing step for Corel-5K and SUN databases. Hence we keep them all to avoid over-filtering. Therefore, in the following experimental analysis, if not specified otherwise, only the visual modality gets filtered.

\textbf{Test set likelihood.} It is used for evaluating the generalization performance of a topic model. If a model fit the data better, it should assign a higher likelihood to the test set. We compute the per-image average negative log-likelihood of the test set with various hidden topic number $K$ to check 1) the capability of our model to fit the data and 2) the robustness of our model to various $K$.

Fig. \ref{fig:testsetlikelihood} illustrates the results conducted on Corel-5K. As expected, VSEC-LDA provides a better fit than Corr-LDA on the test data regardless of the initial vocabulary size. Also, the larger the vocabulary size is, the larger the negative log likelihood is for Corr-LDA. This implies that a dense vocabulary is beneficial for learning the model. On the other hand, our approach achieves similar optimal likelihood across different settings of the initial size, and it is lower than that of Corr-LDA. This phenomenon implies our method can achieve a reasonable vocabulary size robustly. For Corel 5K, when the visual vocabulary size is 1200, the optimal average negative log-likelihood given by both method are similar. However, Corr-LDA fails to fit the data well for the other two vocabulary size settings. In this sense, 1200 is the optimal visual vocabulary size for Corel-5K.

\textbf{Image annotation performance.} After a model is trained, given an image without its captions, one can use the model to compute a distribution over textual word conditioned on the visual word, given by $p(w|v)$. This distribution reflects a prediction of the missing tag for the image.

We show some examples of image annotation in the supplementary material. But if a predicted caption is accurate may be subjective, we employ "perplexity" as an objective measure. Perplexity is introduced from conventional language modeling to evaluate the uncertainty of the predicted captions. It is equivalent algebraically to the inverse of of the geometric mean per-word likelihood. Similar to negative log likelihood measurement, a lower perplexity score indicates a better predictive performance.
\begin{equation}
    perplexity = exp\big(- \frac{\sum_{d}^{D}\sum_{n = 1}^{M_d}log p(w_n|\bold{v}_d)}{\sum_{d=1}^{D}M_d}\big)
\end{equation}
\begin{figure*}
    \centering
    \includegraphics[width=0.32\linewidth,height = 3cm]{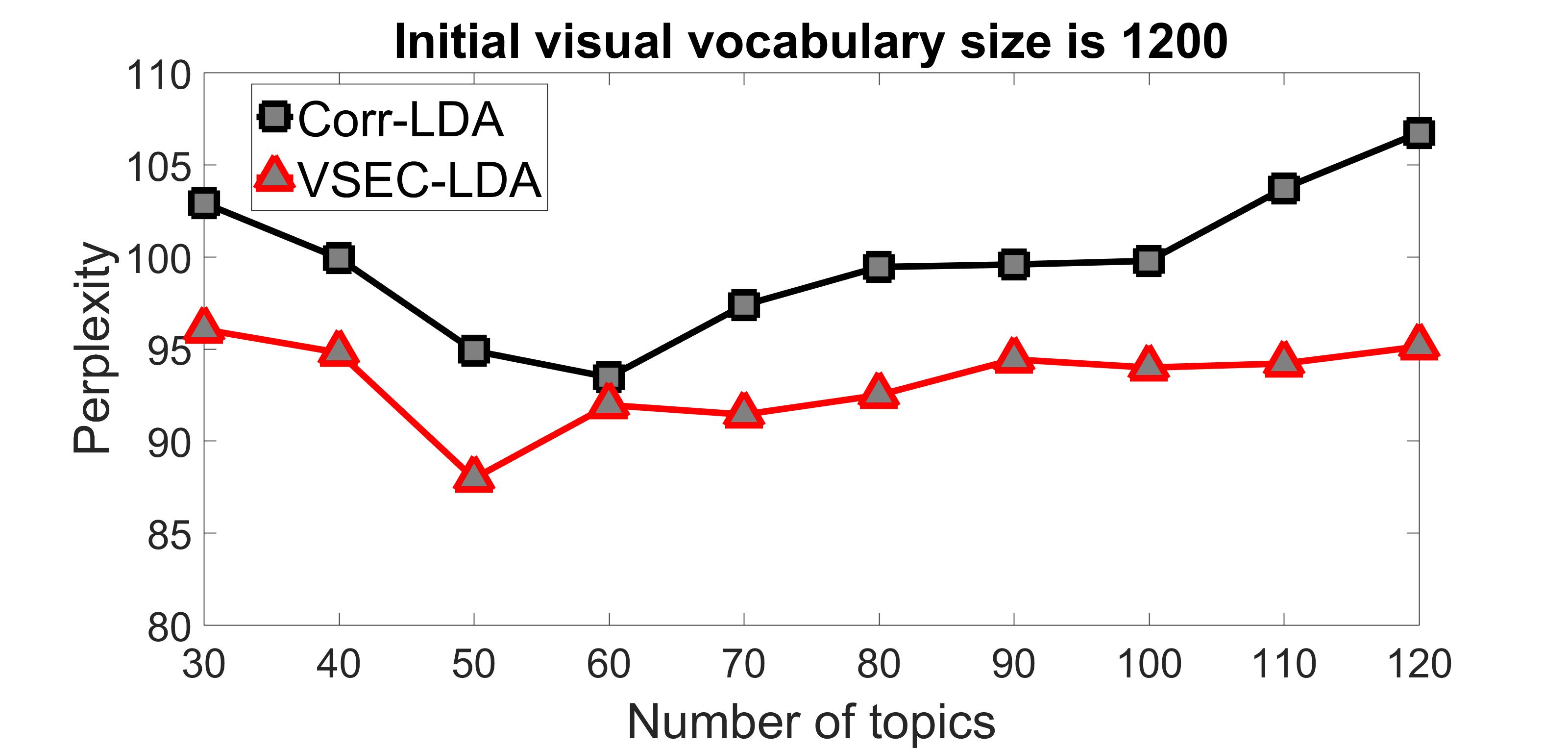}
    \includegraphics[width=0.32\linewidth,height = 3cm]{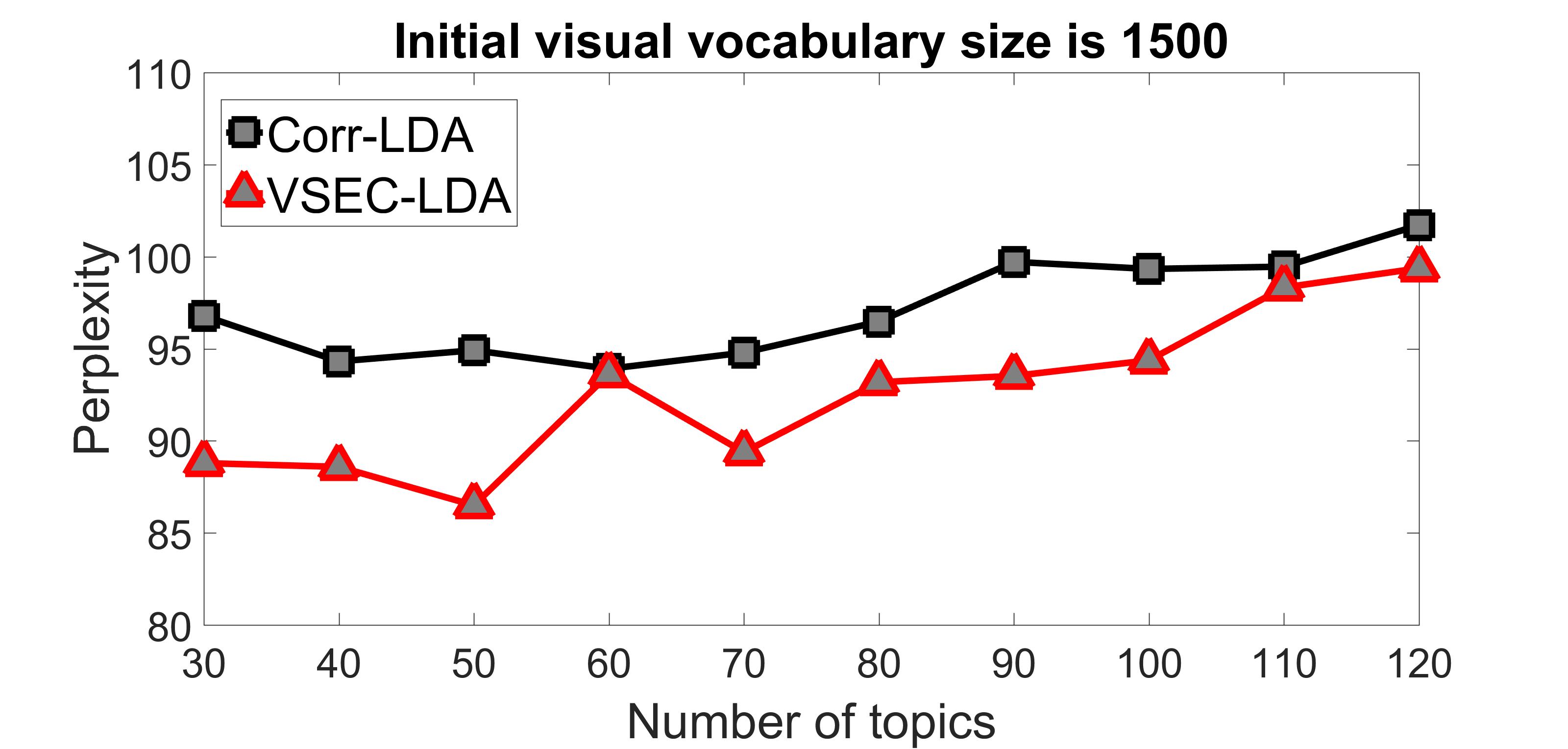}
    \includegraphics[width=0.32\linewidth,height = 3cm]{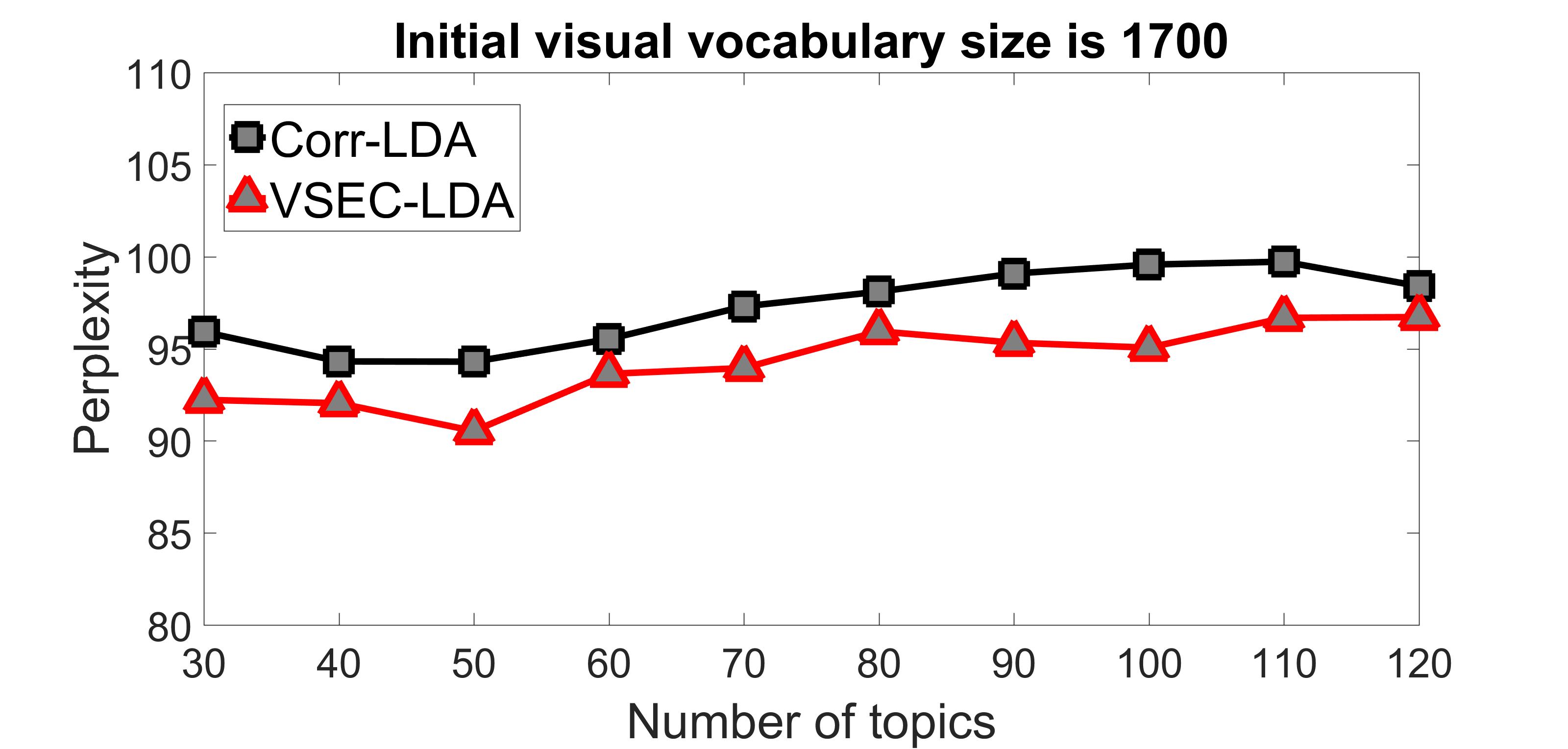}
    \caption{Predictive perplexity on Corel-5K along with different number of topics $k$ of Corr-LDA and VSEC-LDA. From left to right, the visual vocabulary size ranges from 1200 to 1700. The number of topics $k$ ranges from 30 to 120. For most cases, both methods achieve their own optimal solution when the number of topics $k$ is around 50.} 
    \label{fig:caption_perplexity}
\end{figure*}

Fig. \ref{fig:caption_perplexity} shows the predictive perplexity of Corel-5K (similar trend for SUN) with various topics $k$, where we notice that our model always achieves a lower value than Corr-LDA. It is worth mentioning that, with the initial visual vocabulary size increasing (from left to right in Fig. \ref{fig:caption_perplexity}), both methods gradually reach lower perplexity value, and the difference between the methods decreases. This is because when the number of topics $k$ increases, words close to each other (in terms of having similar probability given the topic-word distribution under the same topic) will have a higher chance of being assigned to different topics. 

\begin{figure}
    \centering
    \includegraphics[width=0.8\linewidth,height = 3.5 cm]{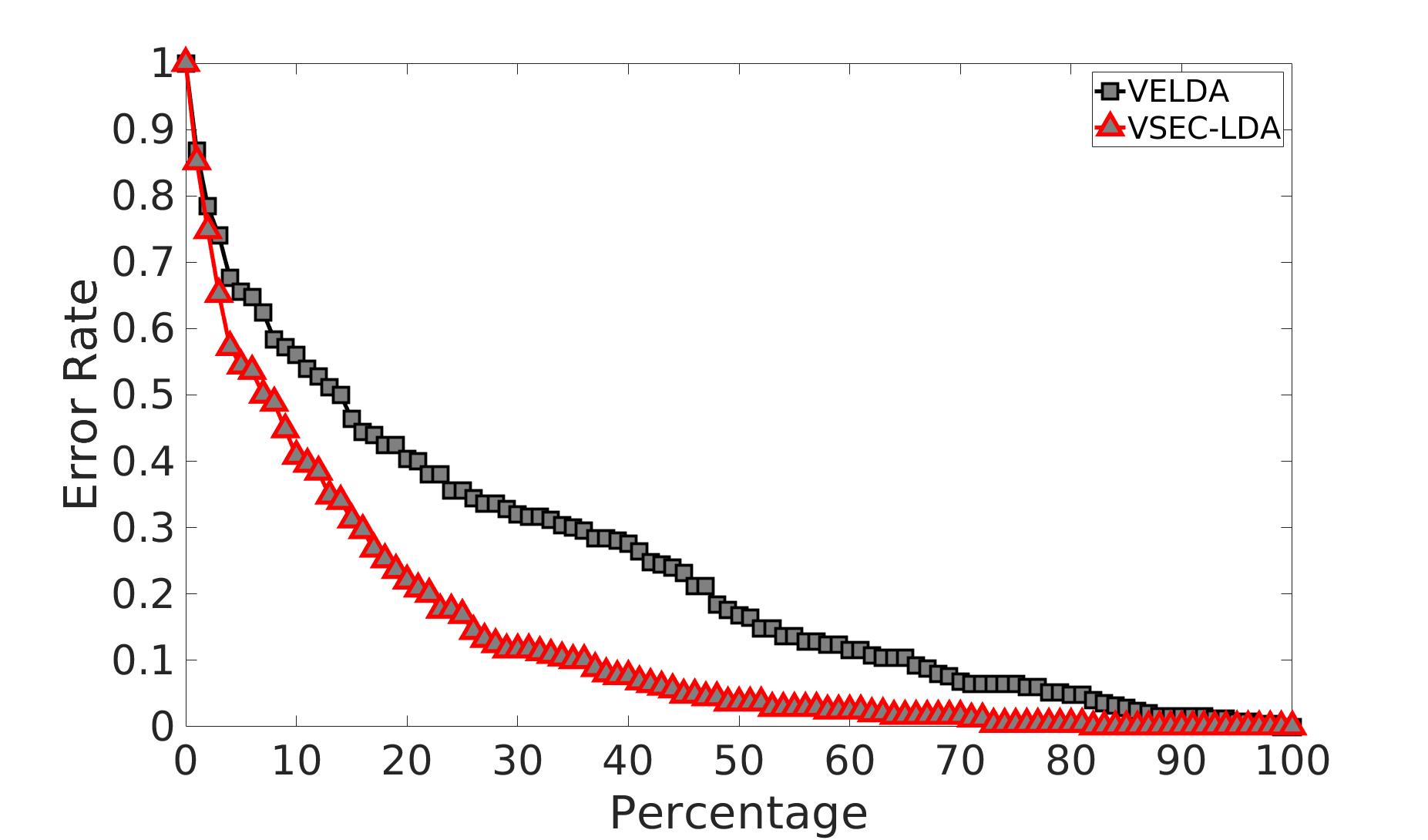}
    \caption{Retrieval error rate by the percentage of POTD database.}
    \label{fig:three_view_error_rate}
\end{figure}

\textbf{Text-based image retrieval.} If a model is learned well, the relationship between visual and textual words should be tight, meaning given one modality, it can refer to the other modality. Thus, we consider text-based image retrieval task: Given the textual word of an image, attempt to retrieve its accompanying image from the image set. To be specific, given a textual query $Q = {1,2,...,q}$, we compute a relevance score for image $i$ by the following formula:
\begin{equation}
  score_{i}=p(T|\theta_i)=\prod_{q = 1}^{Q}p(q|\theta_i)=\prod_{q = 1}^{Q}(\sum_{k = 1}^{K}\theta_{i,k})
\end{equation}
where$p(q|\theta_i)$ is the probability of the $q$-th query word under the distribution $p(q|C)p(C|\theta)$. Having the score of each image at hand, we return a list of images ranked in descending order by conditional likelihood. Since we know the corresponding image for each textual query list, the position of the ground-truth image in the ranked list is considered as a measurement: An image is correctly retrieved if it appears in the top $t$ percent of the image test set, and the error rate is computed by Eq. (\ref{err_rate}).
\begin{equation}
    err\_rate(e) = 1 -  \sum_{i = 1}^{D} (\frac{rank_{i}}{D} <e), e \in [0,1]
    \label{err_rate}
\end{equation}
wherewhere $rank_i$ is the rank of the retrieved image $i$ in the ranking list. Here we only present the results of Model II and VELDA on Wikipedia POTD as an example to show both the performance improvement and demonstrate the generalization ability of our model. More experiment results and analysis of Model I and Corr-LDA with various configurations can be found in the supplemental material. Since the vocabulary size of all the three modalities is around or more than 1000, we consider removing irrelevant words from all modalities. Results of the error rate of POTD are shown in Fig. \ref{fig:three_view_error_rate}. The figure depicts the retrieval errors averaged over all testing queries. A better performance is equivalent to lower error rate：Curves are closer to the bottom left corner. From the curve, it is obvious that our method achieve a much lower error rate than that of VELDA approach. We also compute the recall on the top 10\% level. Model II gets 59.2\% (148/250) of images correctly retrieved in the top 10\% of the ranked list. Compared wit VELDA, our Model II improves retrieval performance by 14.8\% or 5.6\% if compared with the results reported in the original paper.

\begin{table}
\begin{tabular}{ ||l|l|| }
\hline
\multicolumn{2}{|c|}{topic28}\\
\hline
\multirow{3}{*}{VSEC-LDA} & sentences, grammatical, sentence \\
& attended, touch, sounds, listening  \\
& sentence\_comprehension, listened, vision  \\
\hline
\multirow{3}{*}{GC-LDA} & sentences, sentence, comprehension \\
&  semantic, input, slowly  \\
& modality, touch, listening, melody \\
\hline

\end{tabular}
\caption{Examples of top-10 textual words retrieved by one topic. (More examples in Supplemental Materials.)}
\label{table_topic_word}
\end{table}

\textbf{Inspecting learned topics for the Neorosynth database} In Table \ref{table_topic_word}, we illustrate the top 10 words according to the topic distributions learned by Model III and GC-LDA respectively for a sample topic (topic28). Here there is no groundtruth for direct evaluation as before, and thus we rely on examining how relevant the words are. By looking at how the textual words are related to each topic, we may identify the function of the activated area in the brain. In the table, the second and third rows show the words from topic28 by VSEC-LDA and GC-LDA respectively. It is obvious the words given by our model are all related to sentence understanding, while GC-LDA  gives words like `slowly' and `input', which are not related to the given topic. 

\section{Conclusion}

We proposed VSEC-LDA, a new topic model with integrated vocabulary selection. Existing topic models generally require a data pre-processing step before learning the models. This step is typically independent of the model learning step and thus does not necessarily support finding an optimal topic model for a given problem. The proposed approach uses entropy-based metrics to account for the relevant importance of the words to the model and dynamically determine the relevant words while learning the model. Experiments with synthetic databases show that the proposed model can select the relevant words effectively. Real-data-based experiments also demonstrate improved performance of the proposed model in different applications.

{\small
\bibliographystyle{ieee_fullname}
\bibliography{egbib}
}

\section*{Supplementary material}
\section{Generative process for the other two variants of VESC-LDA}
For completeness, we present the generative process for the other two variants here. Only the steps that are different from Model I are shown here for brevity. The Model II, which has 3 modalities in total, the generative process for each visual word in both visual modalities is the same as Model I. Note the two visual modalities are independent of each other. The generative process for each textual word is summarized as following:
\begin{enumerate}
\item Sample a relevance distribution $\lambda$ from $Dir(\eta)$
\item For each relevant textual word $w_n$, $n\in$ \{\ $1,...,N$ \}\
    \begin{enumerate}
    \item Sample relevance type $r_m$ from $Mult(\lambda_{w_n})$
    \item If $r_m=0$, sample topic indicator variable $y_n$ from Multinomial( $\frac{m_{d}^{1}}{m_d},...,\frac{m_{d}^{k}}{m_d},...,\frac{m_{d}^{K}}{m_d}$), where $m_{d}^{k}$ represents the number of times that $k$-th topic from the first visual modality appear in current document $d$.
    \item If $r_m = 1$, repeat previous step, but consider the topic from the second visual modality.
    \item Sample the textual word $w_n$ from Multinomial ($\psi_{y_n}$).
    \end{enumerate}
\item For each irrelevant textual word $w_{in}$, ${in}\in$ \{\ $1,...,{IN}_{d}$ \}\
    \begin{enumerate}
    \item Sample the textual word $w_{in}$ from Uniform( $1,...,IN_{d}$).
\end{enumerate}
\end{enumerate}
Model III presented in our paper has a subregion-topic distribution with $R$ components. So for each visual word, it first draws a subregion indicator from multinomial distribution, then samples the visual word from corresponding topic distribution. The more clear illustration of the model is shown in Fig. \ref{fig:model_structure}.
\begin{figure*}
    \centering
    \includegraphics[width=0.9\linewidth,height = 9cm]{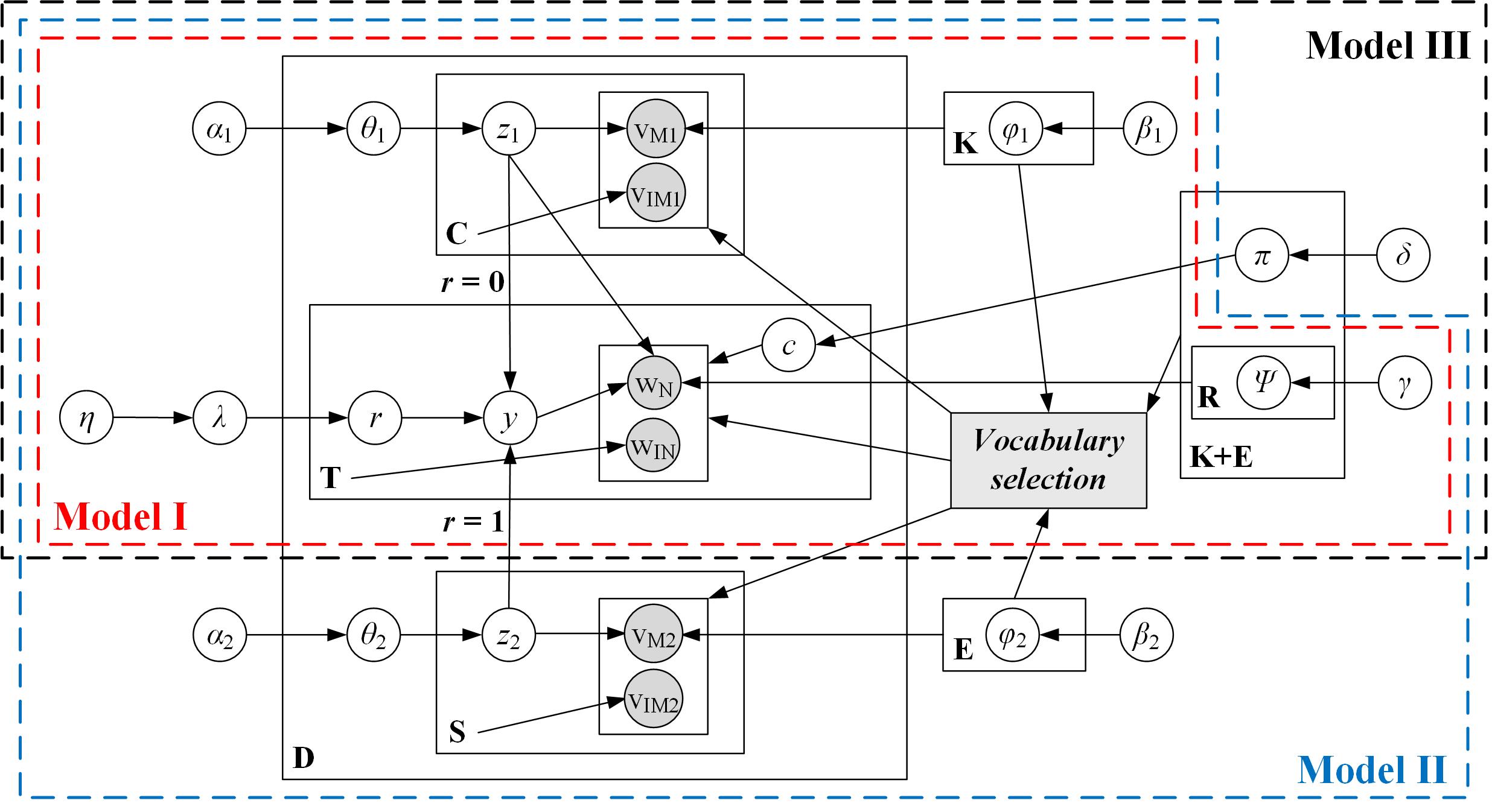}
    \caption{ The generative model in VSEC-LDA. The area within the red dash line is Model I (the basic model). The area within the blue dash line represents Model II, which deals with multiply modalities. The area within the black dash line is Model III, targeting neuro-image analysis.}
    \label{fig:model_structure}
\end{figure*}

\begin{table*}
\begin{subfigure}
    \centering
\includegraphics[width= 0.33\linewidth, height = 2.5cm]{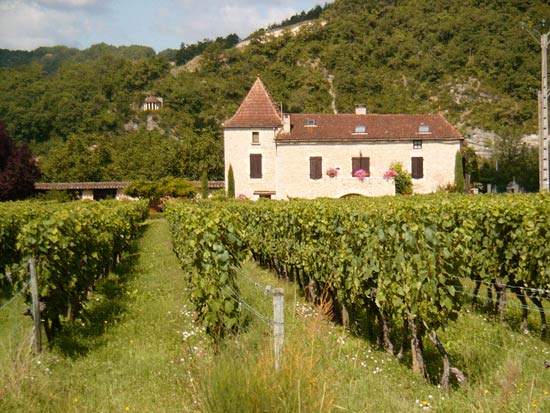}
\end{subfigure}
\begin{subfigure}
    \centering
\includegraphics[width=0.33\linewidth, height = 2.5cm ]{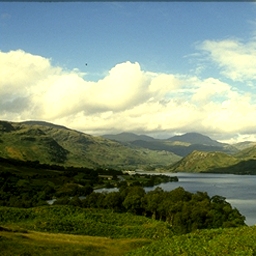}
\end{subfigure}
\begin{subfigure}
    \centering
\includegraphics[width=0.33\linewidth, height = 2.5cm]{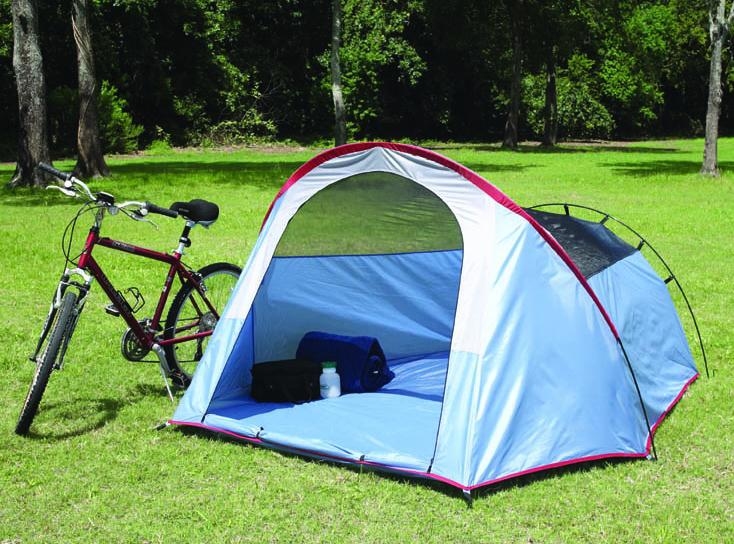}
\end{subfigure}
\begin{tabular}{p{0.64\columnwidth}}
\textbf{Groudtruth:}\\ house building mountain \\pole vineyard window trees\\
\textbf{VSEC-LDA:}\\ house tree mountain grass window\\
\textbf{Corr-LDA:}\\ building trees window
\end{tabular}
\begin{tabular}{p{0.64\columnwidth}}
\textbf{Groudtruth:}\\ sky mountain ground \\grass river water tree \\
\textbf{VSEC-LDA:}\\ sky cloud mountains grass water\\
\textbf{Corr-LDA:}\\ sky mountain trees\\
\end{tabular}
\begin{tabular}{p{0.64\columnwidth}}
\textbf{Groudtruth:}\\ trees grass bottles bag \\tree truck tent bicycle \\
\textbf{VSEC-LDA:}\\ bicycle grass tent\\
\textbf{Corr-LDA:}\\ tree tent bicycle
\end{tabular}
\caption{Example images from the test set of SUN and their automatic annotations under different models.}
\label{table:image_annotations}
\end{table*}
\begin{figure*}
    \centering
    \includegraphics[width=0.32\linewidth,height = 2.5cm]{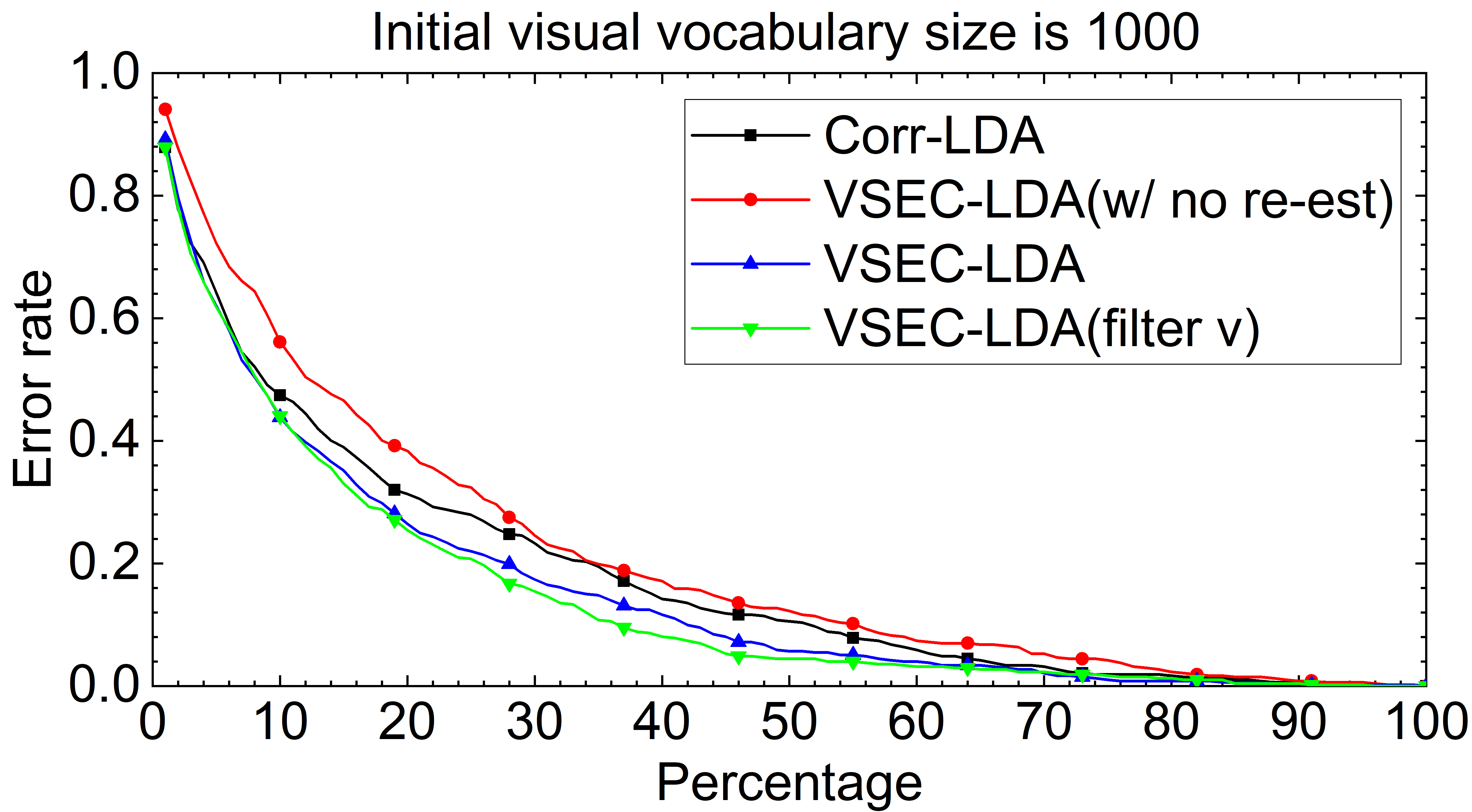}
    \includegraphics[width=0.32\linewidth,height = 2.5cm]{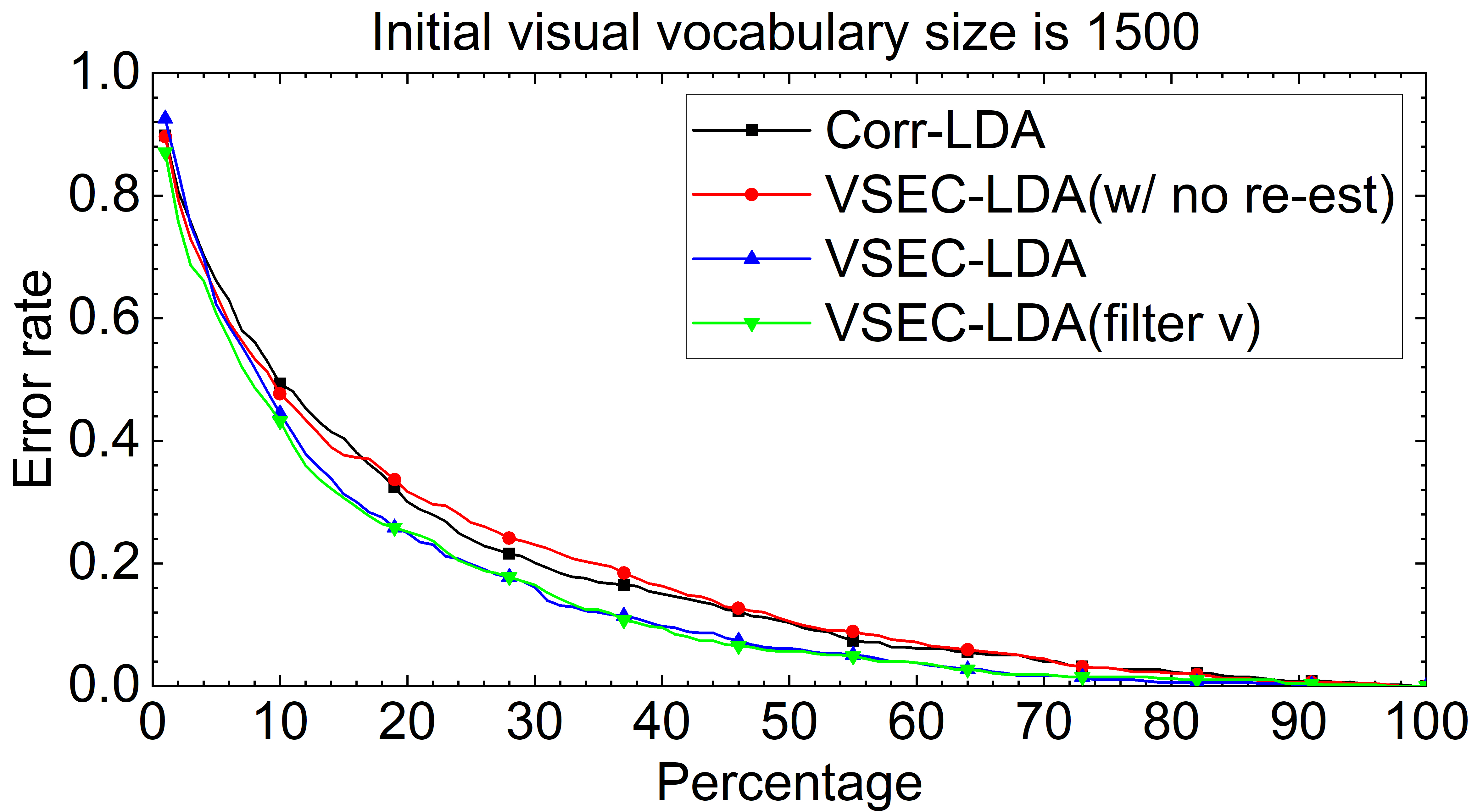}
    \includegraphics[width=0.32\linewidth,height = 2.5cm]{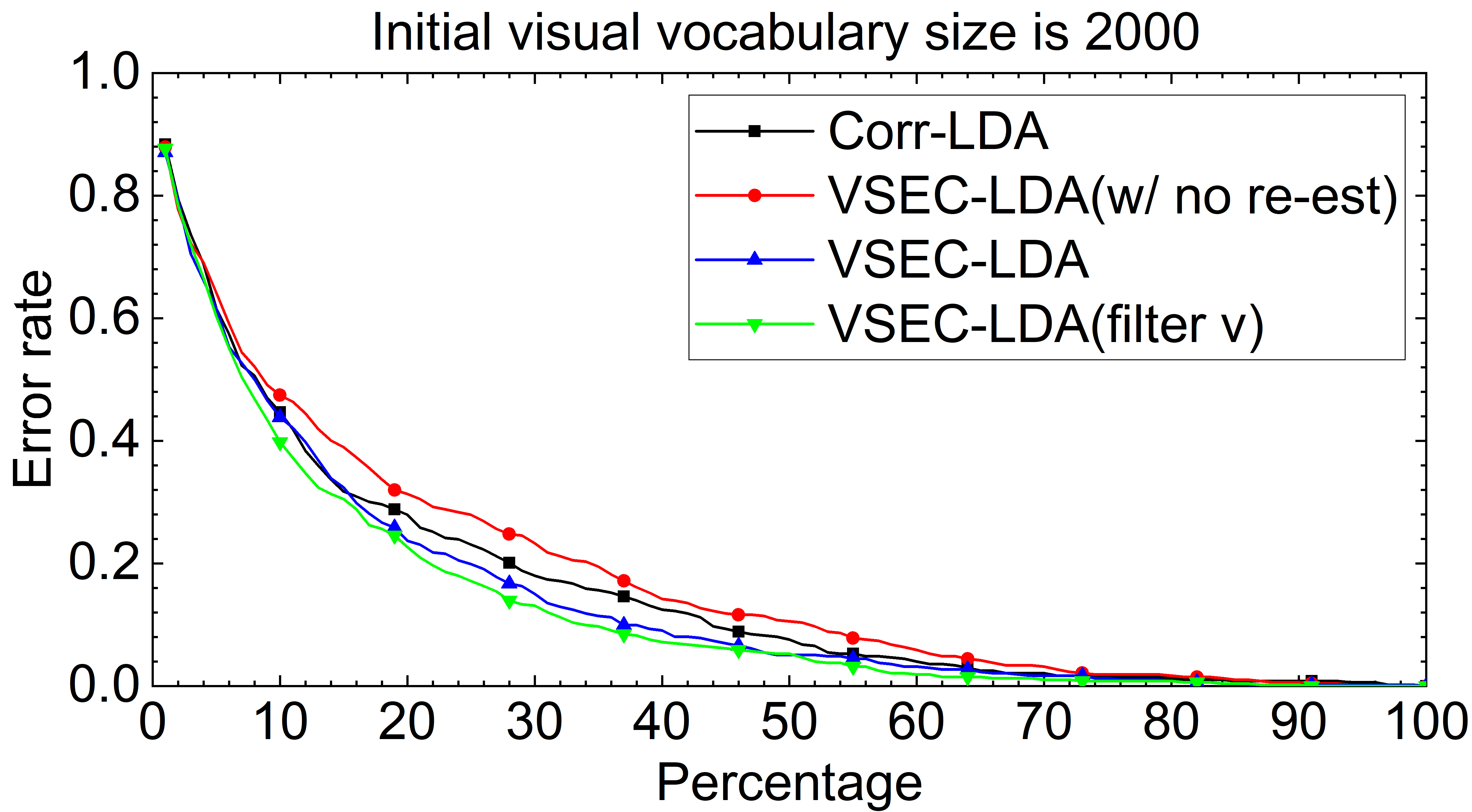}
    \caption{Retrieval error rate by the percentage of the ranked list considered.}
    \label{fig:err_rate}
\end{figure*}
\section{More experiment results}
\begin{figure}
    \centering
    \includegraphics[width=0.8\linewidth, height = 3cm]{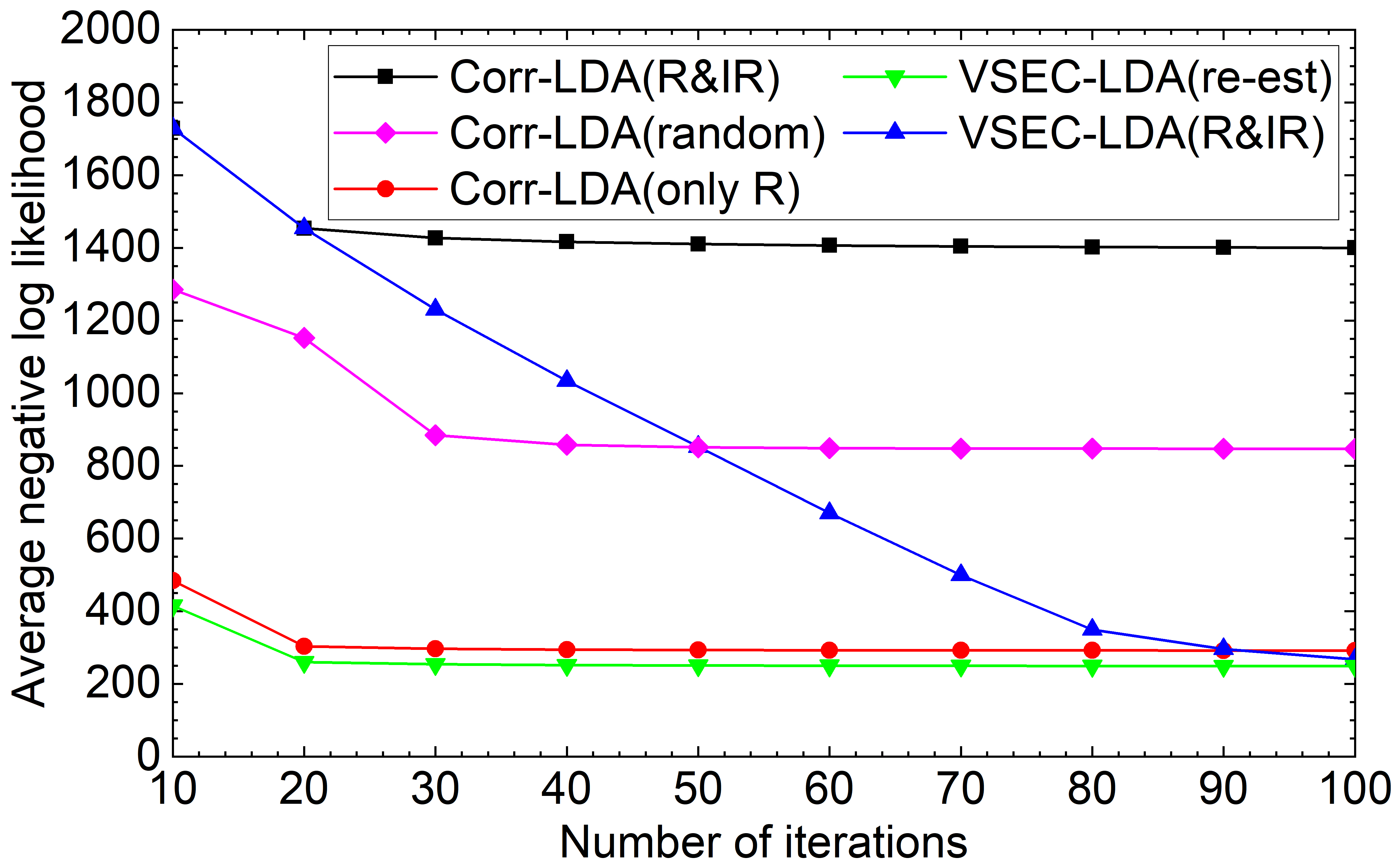}
    \caption{The trend of average negative log likelihood versus with the inference iterations. The corpus is of 80/800 relevant words and 100/1000 irrelevant words. The black curve represents the Corr-LDA performance; the blue curve represents the VSEC-LDA(without re-estimation step);the pink curve represents Corr-LDA conducts on a subset of the original corpus with randomly selected words; the red curve represents Corr-LDA with only the relevant word and the blue one shows VSEC-LDA with re-estimation step.}
    \label{fig:perplexity_syn}
\end{figure}

\begin{table*}[h]
    \centering
    \begin{tabular}{||c||c|c|c|c|c|c||}
    \hline
     \# of irrelevant words& 40/400&60/600 &80/800&100/1000 & 120/1200&140/1400\\\hline
     \# of filtered textual words & 40 $\pm$ 2& 59 $\pm$ 2 & 72 $\pm$ 5 & 86$\pm$ 7 &105 $\pm$ 7 &121 $\pm$ 7\\\hline
     \# of filtered visual words & 392 $\pm$ 9 &587 $\pm$ 11 & 775 $\pm$ 8 & 966 $\pm$ 15 & 1089 $\pm$ 12 & 1121 $\pm$ 15\\ \hline
    \end{tabular}
    \caption{Mean and standard deviation of number of irrelevant words being filtered via 50 rounds experiments. When the number of irrelevant words is greater than 1.5 times of relevant words, it is challenging for the filtering component to do its job within limited iterations.}
    \label{tab:filter_limit2}
\end{table*}
In this section, we will further demonstrate the improved performance of our model compared with baselines by more experiment results with both synthetic database and real database.
\subsection{Experiments with synthetic database}
The full results of testing the limitation of vocabulary selection is shown in Table \ref{tab:filter_limit2}. The maximum training iteration is 150 and fixed filtering threshold is applied for all experiments regarding fair comparison. With the constraint of maximum training iteration, the filtering component starts to see difficulty when irrelevant words become dominant compared to relevant word in the corpus, which is intuitive.

Moreover, Figure \ref{fig:perplexity_syn} illustrates the trend of average negative log-likelihood, which is used for measuring the quality of modeling (the lower the value is, the better the quality is), along with the inference iterations on one training set. The black curve represents the Corr-LDA performs on corpus of both relevant and irrelevant words, while the blue curve represents VSEC-LDA (without re-estimation step) performs on the same corpus. It is obvious that our method gave better results than Corr-LDA. Moreover, if we conduct re-estimation step, VSEC-LDA (green curve) performs even better than Corr-LDA (red curve) on corpus without irrelevant words. We also tried to randomly select words from the given corpus to form a new corpus, and applied Corr-LDA to the new corpus. The result (pink curve) shows it achieves better fitting than Corr-LDA with original corpus, but by simply randomly selecting words, it cannot get optimal result.

\begin{table}
\begin{tabular}{ ||l|l|| }
\hline
\multicolumn{2}{|c|}{topic28}\\
\hline
\multirow{3}{*}{VSEC-LDA} & sentences, grammatical, sentence \\
& attended, touch, sounds, listening  \\
& sentence\_comprehension, listened, vision  \\
\hline
\multirow{3}{*}{GC-LDA} & sentences, sentence, comprehension \\
&  semantic, input, slowly  \\
& modality, touch, listening, melody \\
\hline
\multicolumn{2}{|c|}{topic57}\\
\hline
\multirow{3}{*}{VSEC-LDA} & movement, finger, timing, imitation \\
 & motor\_control, preparation, adjustment \\
 & response\_selection, tapping, executed \\
\hline
\multirow{3}{*}{GC-LDA}  & motor, movement, finger \\
 & sensory, sensorimotor, visual \\
 & imitation, visually, eight, motor\_control \\
\hline
\end{tabular}
\caption{Examples of top-10 textual words retrieved by two topics, respectively.}
\label{table_topic_word}
\end{table}

\subsection{Experiment with real database}
We show some examples of image annotation from SUN in Table \ref{table:image_annotations}. Both Corr-LDA and our model can capture the majority objects (captions) in the given image. But the retrieved textual words are more correlated given by our model.

We also conduct experiment to test the text-based image retrieval performance of Model I and Corr-LDA on SUN. We consider filtering words on both modalities or only visual modality for SUN, of which textual vocabulary size is 695. Results of the error rate of SUN are shown in Fig. (\ref{fig:err_rate}). Each figure depicts the retrieval errors averaged over all testing queries. For all figures, better performance is equivalent to lower error rate：Curves are closer to the bottom left corner. In general, our method with re-estimation step (red and blue curve) performs the best, followed by the baseline (black curve). Our method without re-estimation step (red curve) performs slightly worse than the baseline. This phenomenon demonstrates performing filter and model parameter estimation simultaneously sometimes leads to a sub-optimal parameter estimation for the model. Therefore, re-estimating the model parameter with filtering component turned off is an intuitive way to achieve optimal estimation. It is worth mentioning that, with filtering component performing on both the modalities or one of them, our method performed differently, and the best performance achieved when the filtering component only conducted the visual modality. This is not surprising since the words we used to build our vocabulary are not the raw tags/objects, and have been pre-processed. Moreover, we found regardless the difference initial conditions, we got similar relevant textual words when the model converged, only around 30 words being filtered. Thus if excrescent filtering is applied, the performance will depreciate.

Table \ref{table_topic_word} shows the top 10 words according to the topic distributions learned by Model III and GC-LDA respectively for sample topics (topic28 and topic 57). Here there is no groundtruth for direct evaluation as before, and thus we rely on examining how relevant the words are. By looking at how the textual words are related to each topic, we may identify the function of the activated area in the brain. In the table, the second and third rows show the words from topic28 by VSEC-LDA and GC-LDA respectively. It is obvious the words given by our model are all related to sentence understanding, while GC-LDA  gives words like `slowly' and `input', which are not related to the given topic. Similar to topic 28, in topic 57, the topic-word distribution learned by GC-LDA produce word 'eight', which is not related to any of other words.

\end{document}